\pdfoutput=1
\documentclass[11pt]{article}
\usepackage[final]{acl}

\usepackage{times}
\usepackage{latexsym}
\usepackage[T1]{fontenc}
\usepackage[utf8]{inputenc}
\usepackage{microtype}
\usepackage{paralist}
\usepackage{inconsolata}
\usepackage{fdsymbol}
\usepackage{graphicx}
\usepackage{amsmath}
\usepackage{caption}
\usepackage{subcaption}
\usepackage{wrapfig}
\usepackage{svg}
\usepackage[utf8]{inputenc} % allow utf-8 input
\usepackage[T1]{fontenc}    % use 8-bit T1 fonts
\usepackage{hyperref}       % hyperlinks
\usepackage{url}            % simple URL typesetting
\usepackage{booktabs}       % professional-quality tables
\usepackage{arydshln}
\usepackage{colortbl} % For row colors
\usepackage{amsfonts}       % blackboard math symbols
\usepackage{nicefrac}       % compact symbols for 1/2, etc.
\usepackage{microtype}      % microtypography
\usepackage{xcolor}         % colors
\usepackage{algorithm}
\usepackage[noend]{algpseudocode}
\usepackage{array}
\usepackage{tcolorbox}
\usepackage{mathtools}

\newcommand{\minisection}[1]{\noindent{\textbf{#1}}}

\newcommand{\tool}{\textsc{Open-RAG}~}
\newcommand{\toolnospace}{\textsc{Open-RAG}}
\newcommand{\opencrag}{\textsc{Open-CRAG}~}

\newcommand{\modelsevenB}{$_{\textsc{7B}}$}
\newcommand{\modeleightB}{$_{\textsc{8B}}$}
\newcommand{\modelmoesevenB}{$_{\textsc{7B+8}\times\textsc{135M}}$}
\newcommand{\modelmoethirteenB}{$_{\textsc{13B+8}\times\textsc{213M}}$}

\newcommand{\modelthirteenB}{$_{\textsc{13B}}$}
\newcommand{\modelsixfiveB}{$_{\textsc{65B}}$}
\newcommand{\modeloneofourB}{$_{\textsc{104B}}$}

\newcommand{\llm}{$\mathcal{M}_{G}$}

\newcommand{\reltok}{\emph{Relevance}}
\newcommand{\gndtok}{\emph{Grounding}}
\newcommand{\utiltok}{\emph{Utility}}

\title{\toolnospace: Enhanced Retrieval-Augmented Reasoning with Open-Source Large Language Models}

\author{
Shayekh Bin Islam\thanks{\ \ Equal contribution.}$^{,1, 6, 7}$, Md Asib Rahman\footnotemark[1]$^{,1}$, K S M Tozammel Hossain$^{2}$ \\ \bf \ Enamul Hoque$^{3}$, \ Shafiq Joty$^{4}$, \ Md Rizwan Parvez$^{5}$ \\
$^1$Bangladesh University of Engineering and Technology, $^{2}$University of North Texas \\
$^3$York University, Canada, 
$^4$Salesforce Research, 
$^5$Qatar Computing Research Institute (QCRI) \\
$^6$Fatima Al-Fihri Predoctoral Fellowship,
$^7$Cohere For AI Community \\
\texttt{ 
shayekh.bin.islam@gmail.com,
mparvez@hbku.edu.qa }
}

\newcommand{\tabmain}{

\begin{table*}[ht]
    \centering
    \footnotesize
      \setlength{\tabcolsep}{3pt} % final
\resizebox{\textwidth}{!}{
    \begin{tabular}{l  |rrr |rrrr |rrrrrrr} 
        \toprule
          & \multicolumn{3}{c|}{Short-form } & \multicolumn{4}{c|}{Long-form generations} & \multicolumn{6}{c}{Multi-hop generations} & \\
          &  Pop & TQA& Pub & Bio& \multicolumn{3}{c|}{ALCE-ASQA} &  \multicolumn{2}{c}{Hotpot} &  \multicolumn{2}{c}{MuSiQue}& \multicolumn{2}{c}{2WikiMH} & \\
          LM & Acc &Acc& Acc &FS& SM& rg& mau& EM& F1& EM& F1& EM&F1 \\ 
          \midrule
          \multicolumn{14}{c}{\it LMs with proprietary data/retriever} \\
           \midrule 
          Perplexity.ai & --&--&--&71.2 &--&--&--&--&--&--&--&--&-- \\
          RAG 2.0  & -- & -- & -- & -- & -- & -- & -- &  54.0 & -- & -- & -- & -- & --  \\
          ChatGPT & 29.3 &  \textcolor{gray}{\bf 74.3} & 70.1 & 71.8 & 35.3  & 36.2  & 68.8 &  22.4 & 30.0 & 3.1 & 7.3 & 18.7 & 21.7 \\
        
         RAG-ChatGPT &  50.8 & 65.7 &  54.7 & -- & \textcolor{gray}{\bf 40.7} &   \textcolor{gray}{\bf 39.9} & { 79.7} & 55.3 & 69.9 & 31.2 & 43.5  & 44.7  &  54.8  \\

         RAG-Command R+$^*$\modeloneofourB &  \textcolor{gray}{\bf 59.9}  & 74.0 & 46.3 & \textcolor{gray}{\bf 84.0} & -- & -- &--   & 
         60.0
         & 75.8 & 41.3 &  55.4 & \ 57.1 &  66.1 \\
         \cdashline{1-14}
         RQ-RAG$^\dag$\modelsevenB~(ToT) & 57.1 & -- & -- & -- & -- & -- & -- &  62.6 & -- &  41.7 & -- & 44.8 & --  \\
          \midrule  
          \multicolumn{14}{c}{\it Baselines without retrieval} \\
          \midrule 
         Llama2\modelsevenB &  14.7 &30.5&  34.2 &44.5&  7.9&  15.3&  19.0&  3.8&  9.3&  2.0&  3.3& 8.0&14.5 \\ 
         Alpaca\modelsevenB &  23.6 & 54.5  & 49.8 & 45.8 & 18.8 & 29.4 & 61.7 & 4.7 & 11.5 & 2.5 & 3.8 & 15.3 & 20.0  \\
         SAIL\modelsevenB &  22.8 & --  & -- & -- & -- & -- & -- & -- & -- & -- & -- & -- & --  \\
         \cdashline{1-14}
         Llama2\modelthirteenB & 14.7 & 38.5  & 29.4 & 53.4 & 7.2 & 12.4 & 16.0  & 14.9 & 21.6 & 1.3 & 5.4 & 21.4 & 25.2 \\
         Alpaca\modelthirteenB & 24.4  &  61.3  & 55.5  & 50.2 & 22.9 &  32.0&  70.6 & 0.7 & 6.1 & 0.0 & 3.3 & 3.1 & 12.0 \\
         CoVE\modelsixfiveB &  -- & --  & -- & 71.2 & -- & -- & -- & -- & -- & -- & -- & -- & --  \\
          \midrule
         \multicolumn{14}{c}{\it Baselines with retrieval} \\    
          \midrule 
         Llama2\modelsevenB &  38.2 &48.8&  30.0 &78.0&  15.2&  22.1&  32.0&  5.9&  19.4&  3.4&  10.5& 11.9&19.2 \\ 
         Alpaca\modelsevenB  & 46.7 & 64.1  & 40.2 & 76.6  & 30.9 & 33.3 & 57.9 & 23.0 & 35.6 & 6.4 & 14.8 & 18.2 & 23.8 \\
         
         SAIL\modelsevenB &  44.0 & --  & 69.2 & -- & -- & -- & -- & -- & -- & -- & -- & -- & --  \\
         Self-RAG\modelsevenB &  54.9 &66.1&  72.0 &78.6&  30.2&  35.7&  74.9&  40.2&  54.3&  22.1&  33.2& 24.6&35.8 \\ 
        \cdashline{1-14}
         Llama2\modelthirteenB & 38.2  & 42.5 & 30.0 & 78.0 & 15.2 & 22.1 & 32.0 & 26.7 & 38.5 & 10.8 & 18.6 & 20.2 & 27.4 \\
          Alpaca\modelthirteenB & 46.1& {66.9}  &  51.1 & 77.7 & { 34.8} & 36.7 & 56.6 & 12.3 & 27.3 & 2.6 & 10.7 & 7.0 & 17.1 \\
          Self-RAG\modelthirteenB &  
         56.0 & {67.5} & { 76.3} & 81.1 & 31.6 & 35.9 & 69.7          
         & 44.2 & 58.2 & 22.2 & 40.0 & 17.7 & 31.8  \\ 
         LongChat\modelthirteenB  & -- & -- & -- & -- & -- & -- & -- &  25.0 & 40.6 & 7.9 & 18.9 & 18.2 & 29.2  \\
        \cdashline{1-14}
        
         \rowcolor[gray]{0.9} \toolnospace$^\ddag$\modelmoesevenB
         &  \textbf{58.3} & {\bf 66.3} &  {\bf 75.9} &\textbf{82.2}&  \textbf{31.9}&  \textbf{36.7}&  \textbf{84.3}&  \textbf{63.3}&  \textbf{76.9}&  \textbf{41.6} &  {\bf 55.3} & \textbf{51.5} & \textbf{61.0} \\ 
         \rowcolor[gray]{0.9} \toolnospace\modelmoethirteenB
         &  \textbf{59.5} & {\bf 69.6} &  {\bf 77.2} &\textbf{81.7$^\#$}&  \textbf{36.3}&  \textbf{38.1}&  \textbf{80.0}&  \textbf{66.2}&  \textbf{80.1}&  \textbf{46.0} &  {\bf 60.1} & \textbf{60.7} & \textbf{70.9} \\ 
         
         \bottomrule
    \end{tabular}
}
\caption{
 Model performances on RAG tasks. Pop, TQA, Pub, Bio, Hotpot, MuSiQue, 2WikiMH denote PopQA, TriviaQA, PubHealth, Biography Generations, HotpotQA, MuSiQue-Ans, 2WikiMultihopQA. Acc, FS, SM, rg, mau, EM, and F1 denote accuracy, FactScore (factuality), str-em, rouge (correctness), MAUVE (fluency), exact match, and F1 scores. $^\#$: evaluated using `gpt-3.5-turbo-instruct' instead of `text-davinci-003'.  $^*$: using 4-bit quantized model. $^\dag$: using a proprietary retriever with Tree-of-Thought prompting. $^\ddag$: \tool model with 7.8B total and 7.0B active parameters. 
{\color{gray}Gray} results are best performances with larger/proprietary models.
}
\label{tab:main}
\end{table*}
}

\newcommand{\tabmoeablation}{
\begin{table}[!t]
    \centering
    \resizebox{0.5\textwidth}{!}{
    \begin{tabular}{ccc  | cccc} 
        \toprule
          $N_E$ & $k$ & Epochs & PopQA & PubHealth & \multicolumn{2}{c}{MuSiQue}\\
           & & & Acc & Acc & EM & F1 \\
           \midrule
           \rowcolor[gray]{0.9} 8  & 2 & 1 & {59.8} & 74.6 & 39.6 & 54.4 \\
           16 & 2 & 1 & 59.2 & 74.6 & 40.5 & 54.4 \\
           16 & 4 & 1 & 59.0 & 72.4 & 40.5 & {54.5} \\
        \cdashline{1-7}
           8  & 2 & 2 & 58.3 & {75.9} & {41.6} & 55.3 \\
         
        \bottomrule
    \end{tabular}
    }
    \vspace{-10pt}
    \caption{Ablation study model performances}
    \label{tab:moe_hp}
    \vspace{-3pt}
    
\end{table}

}

\newcommand{\figinference}{
\begin{figure*}[t]
    \vspace{-10pt}
  \begin{center}
    \includegraphics[width=\textwidth]{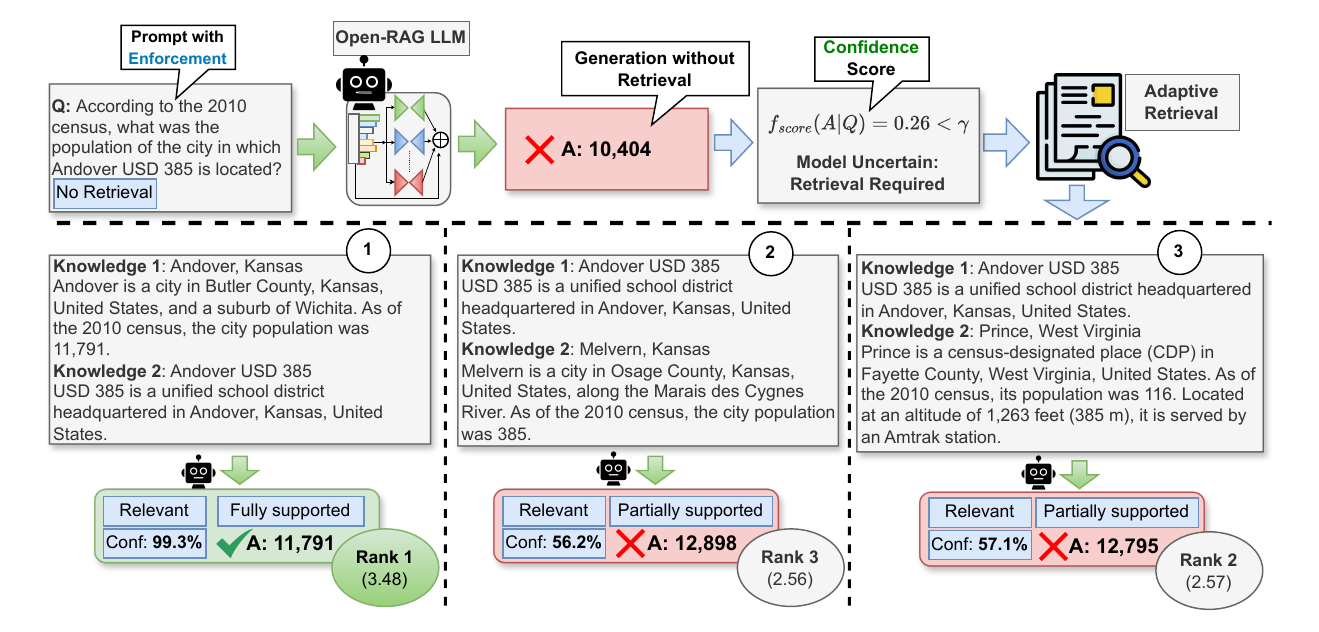}
  \end{center}
  \vspace{-10pt}
  \caption{Inference pipeline in our framework, \toolnospace. It learns to generate retrieval/no\_retrieval tokens, contrasts between relevant and irrelevant contexts, and categorizes answers as partially, fully, or not supported. Then at inference, 
  given a (multi-hop) user query, 
  we first enforce the model to generate an answer with conditional to no\_retrieval as input, and based on the model confidence we dynamically determine if retrieval is needed. 
  } 
  \label{fig:inference-pipeline} 
  \vspace{-10pt}
\end{figure*}
}

\newcommand{\figmoearchitecture}{
    \begin{figure}[h]
      \hspace{-0.45cm}% svg
        \includegraphics[width=0.55\textwidth]{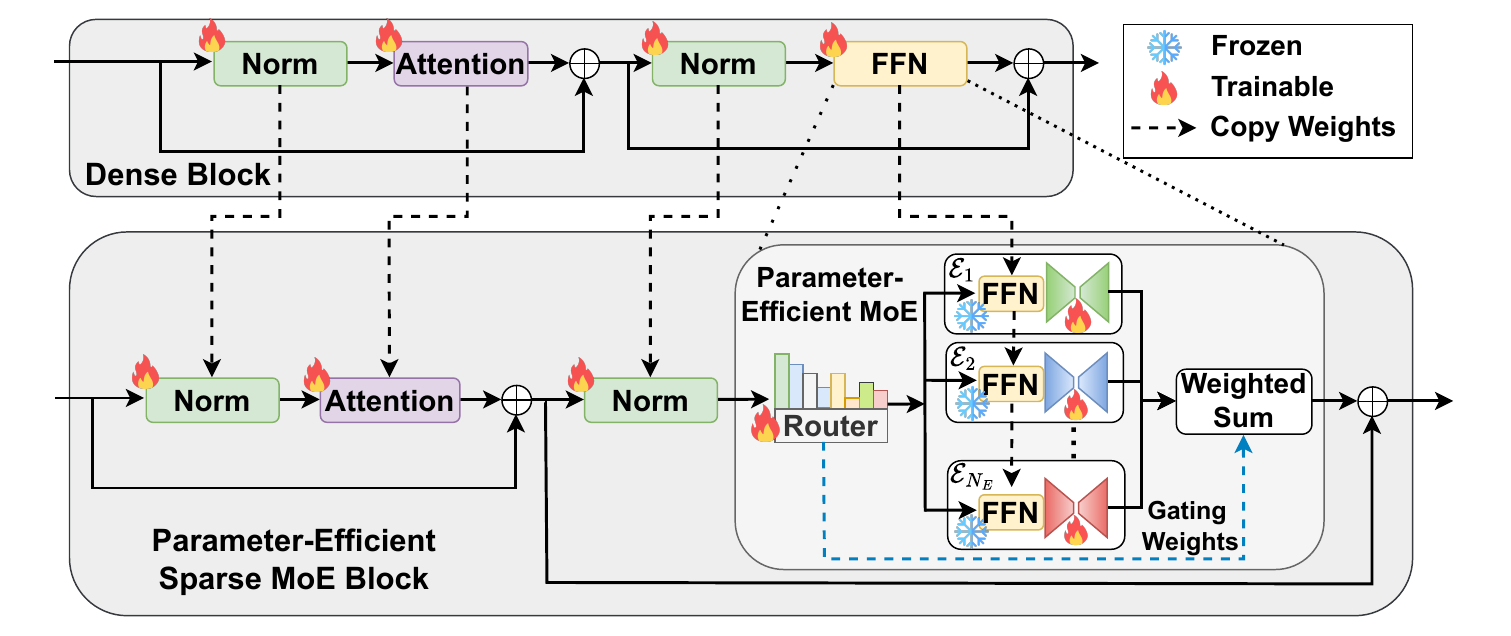}
      \caption{
        Architechture transformation (dense to PEFT MoE) in \toolnospace.  Router $\mathcal{R}$ is trained from scratch. 
        The FFN layer is kept frozen and adapted by parallel-adapter-based experts $\mathbf{E}$. Other layers are being copied.
      }
      \label{fig:moe-architecture}
      \vspace{-10pt}
    \end{figure}
}

\newcommand{\figmhtraindata}{
    \begin{figure*}[t!]
    \vspace{-10pt}
      \begin{center}
        \includegraphics[width=0.98\textwidth]{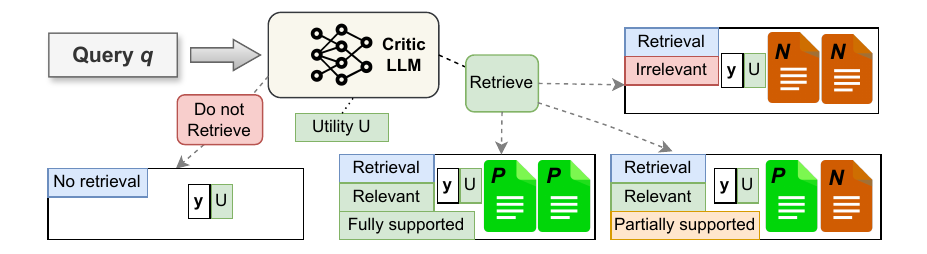}
      \end{center}
      \vspace{-10pt}
      \caption{
      \tool training data preparation involves generating four variations of new training instances from each original pair ($q$, $y$), each incorporating different \emph{reflection} tokens using ground truth/LLM critic and retrieved passages. Our approach enables an LLM not only to reflect on generation quality but also to contrast distractors.}
      \label{fig:mh_data}
      \vspace{-10pt}
    \end{figure*}
}

\newcommand{\figthreshold}{
    \begin{figure*}[h]
    \vspace{-12pt}
    \begin{subfigure}[b]{\textwidth}
        \includegraphics[width=\textwidth]{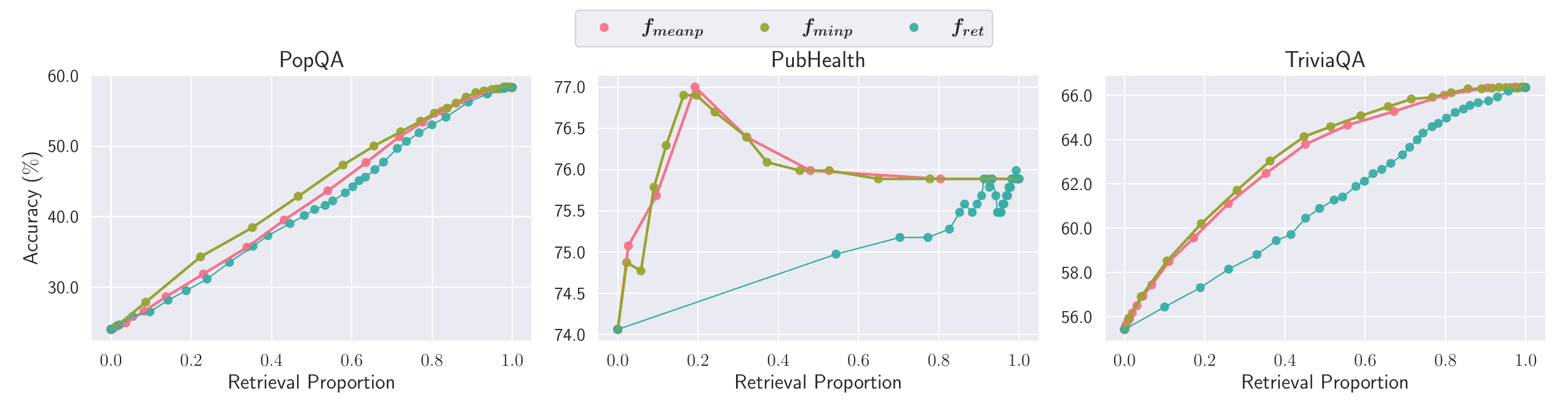}
        \label{fig:thresh_ret}
    \end{subfigure}

    \vspace{-17pt}
    
    \begin{subfigure}[b]{\textwidth}
        \includegraphics[width=\textwidth]{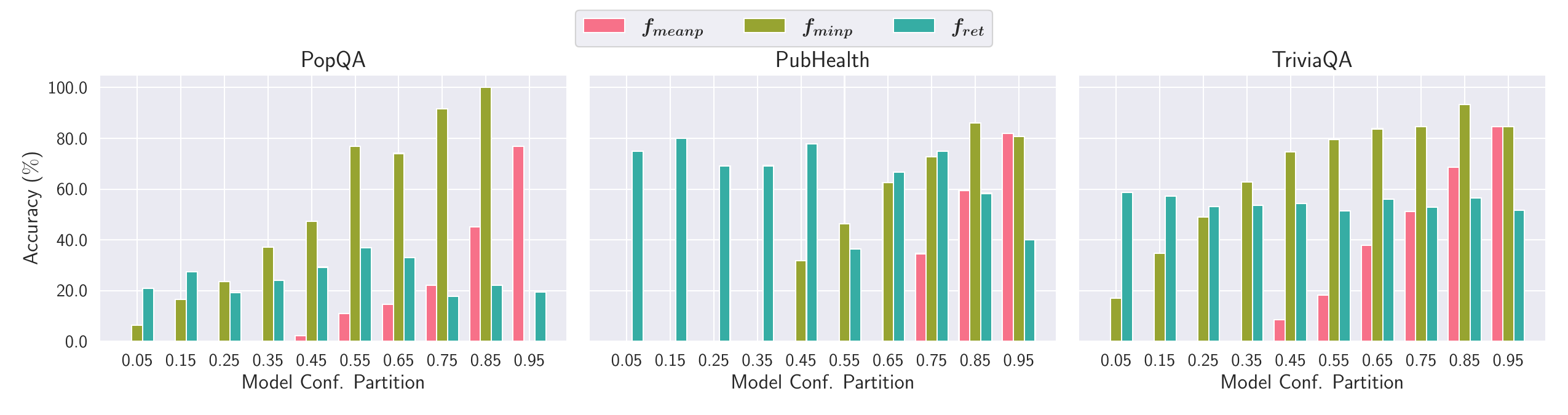}
        \label{fig:thresh_scorebin}
    \end{subfigure}%
    \vspace{-25pt}
    \caption{
    {(Top) Performance vs Retrieval by different adaptive retrieval strategies. }
    (Bottom) Performance vs scores from adaptive retrieval. $f_{ret}$ denotes probability score from external model distilled/predicted \emph{reflection} token.} 
    \vspace{-15pt}
    \label{fig:tradeoff}
    \end{figure*}
}

\newcommand{\figmoedensesparse}{
\begin{figure}[!h]
  \begin{center}
    \includegraphics[width=0.48\textwidth]{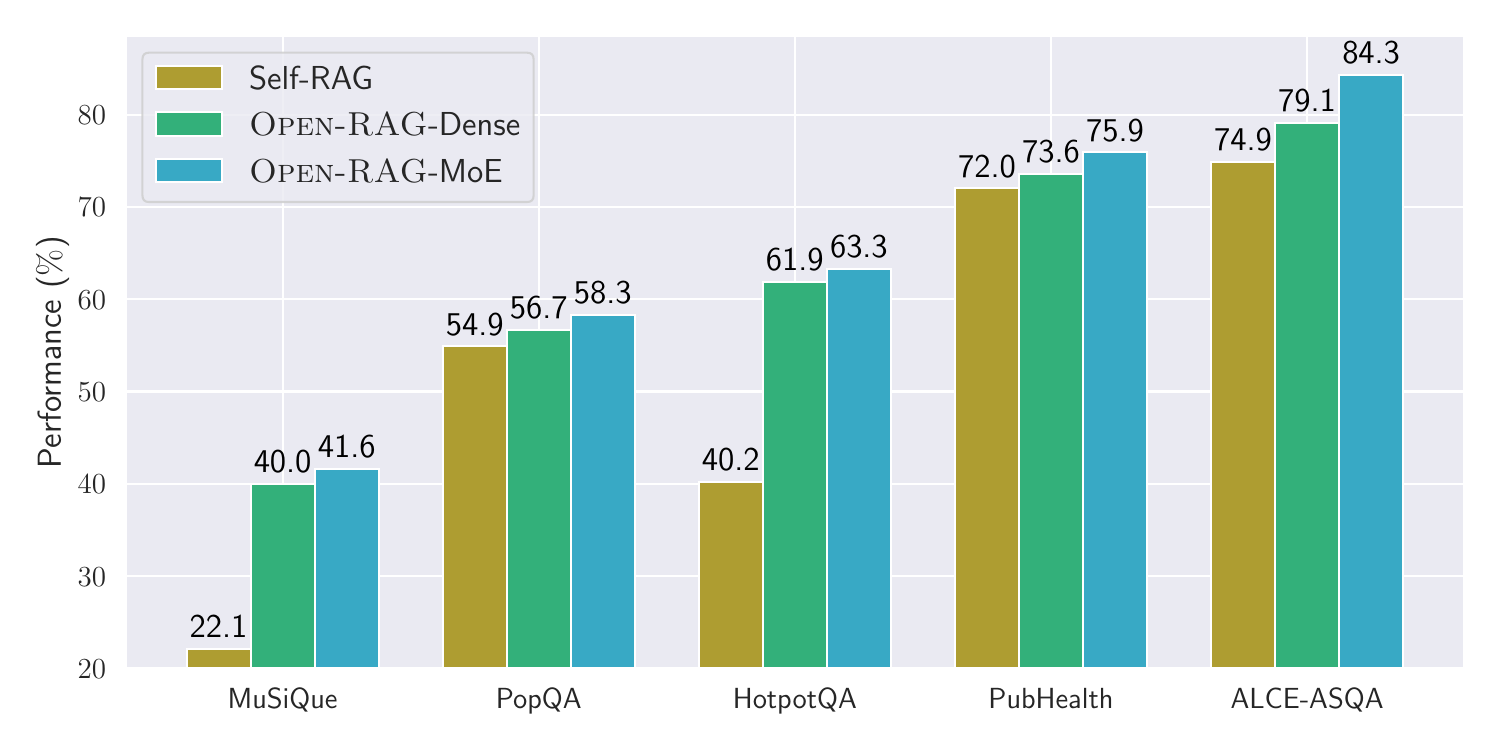}
  \end{center}
  \vspace{-15pt}
  \caption{Performances (MAUVE for ALCE-ASQA; EM for HotpotQA and MuSiQue-Ans; and accuracy for PopQA and PubHealth ) with different architecture. }\label{fig:dense-moe}
\vspace{-15pt}
\end{figure}
}

\newcommand{\figcrag}{
\begin{figure}[h]
\vspace{-5pt}
  \begin{center}
    \includegraphics[width=0.49\textwidth]{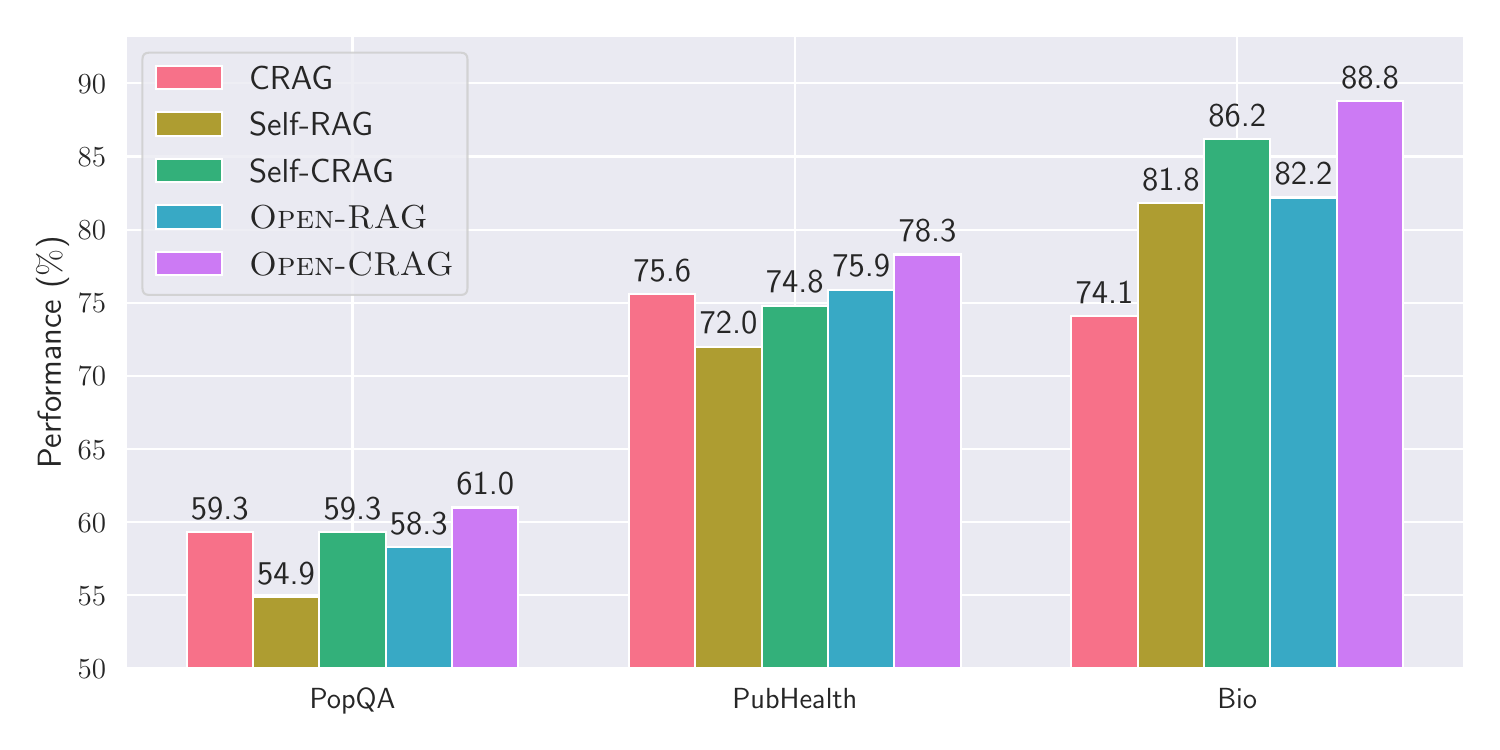}
  \end{center}
  \vspace{-15pt}
  \caption{Model performances utilizing CRAG contexts}\label{fig:crag}
\vspace{-15pt}
\end{figure}
}

\newcommand{\figrouteall}{
    \begin{figure*}[]
      \begin{center}
        \includegraphics[width=\textwidth]{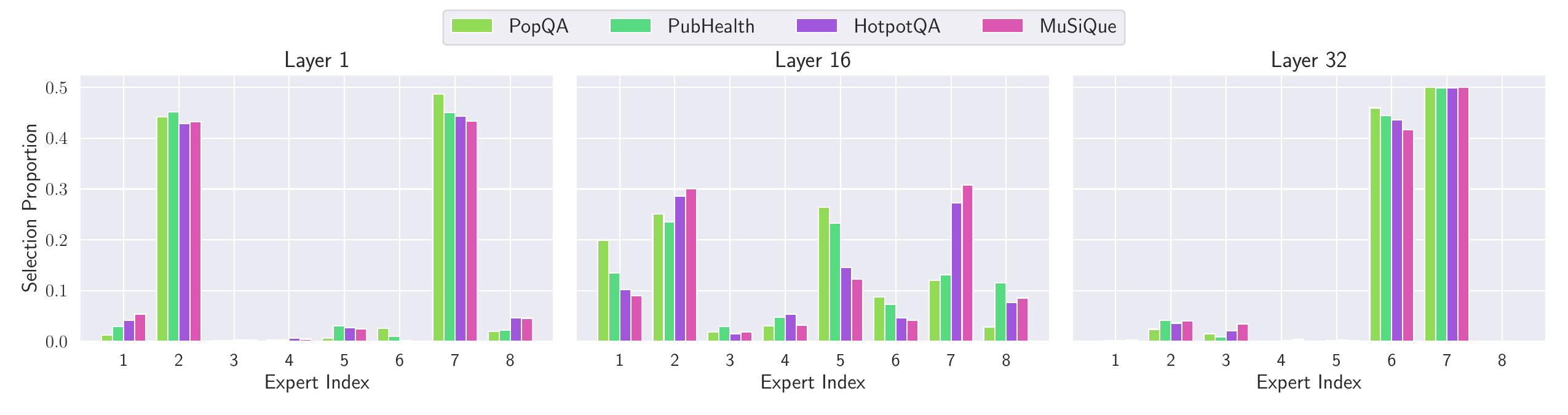} 
      \end{center}
      \vspace{-20pt}
      \caption{Layer-wise expert activation on single-hop (PopQA, PubHealth) vs multi-hop tasks (HotpotQA, MuSiQue).
      }\label{fig:route_all}
      \vspace{-10pt}
    \end{figure*}
}

\begin{document}
\maketitle

% TLDR: An efficient mixture-of-experts based retrieval-augmented LLM with complex reasoning capability

% Keywords: Retrieval-augmented Generation, Mixture-of-Experts, Language Models, Retrieval-augmented LMs

\begin{abstract}

Retrieval-Augmented Generation (RAG) {has been shown to} enhance the factual accuracy of Large Language Models (LLMs), but existing methods often suffer from limited reasoning capabilities in effectively using  the retrieved evidence, particularly when using open-source LLMs. To mitigate this gap, we introduce a novel framework, {\bf\toolnospace}, designed to enhance reasoning capabilities in RAG with open-source LLMs. Our framework transforms an arbitrary dense LLM into a parameter-efficient sparse mixture of experts (MoE) model capable of handling complex reasoning tasks, including both single- and multi-hop queries. \tool uniquely trains the model to navigate challenging distractors that appear relevant but are misleading. 
As a result, 
\tool leverages latent learning, dynamically selecting relevant experts and integrating external knowledge effectively for more accurate and contextually relevant responses. 
In addition, 
we propose a hybrid adaptive retrieval method to determine retrieval necessity and balance the trade-off between performance gain and inference speed. Experimental results show that the Llama2-7B-based \tool outperforms state-of-the-art LLMs and RAG models 
such as 
ChatGPT, Self-RAG, and Command R+ 
in various knowledge-intensive tasks. 
We open-source our code and models at \url{https://openragmoe.github.io/}

\end{abstract}

\section{Introduction}

The rapid advancement of Large Language Models (LLMs) has significantly improved various  NLP tasks \cite{open-llm-leaderboard}. However, these models often suffer from factual inaccuracies \cite{min-etal-2023-factscore, mallen2022not}. Retrieval-Augmented Generation (RAG) has emerged as a promising approach to integrate LLMs with external knowledge, thereby improving generation accuracy \cite{asai-etal-2023-retrieval-akari-application, lewis2020retrieval}. Despite this, existing RAG methods demonstrate limited reasoning capabilities, particularly when employing open-source LLMs and addressing high-complexity queries such as multi-hop retrieval augmented tasks \cite{jeong2024adaptive-rag, zhang2024raft}. Thus, building an effective RAG model using open-source LLMs remains an open challenge. To address this gap, we present {\bf \toolnospace}, a novel framework aimed at improving reasoning capabilities in RAG with open-source LLMs.

Reasoning over retrieved documents is particularly difficult. In general, retrievers are imperfect and can return noisy passages \cite{shi2023large}. The generated outputs can also be inconsistent with retrieved passages \cite{gao-etal-2023-enabling} or can even override the LLM's accurate parametric knowledge \cite{parvez2024evidence}. Approaches like re-ranking or filtering retrieved documents \citep{xu2023recomp, nogueira2020passage, wang2018r} and active retrieval methods (i.e., retrieve only when needed) \citep{mallen2023trust, jiang2023active, trivedi2023interleaving} have shown promising success in tackling these, but they require substantial human annotations, can filter out useful information, often perform sequential and repetitive calls (hence slow), and can still suffer from distracting content, even in relevant passages \cite{filco}.

\figinference

To address and control these behaviors such as retrieval frequency of the RAG model and guide the generation to be contextually consistent, Self-RAG and its variants \cite{selfrag, yan2024correctivecrag, self-bio-rag} adopt a self-reflection-based method. During training, these models learn to generate both task output and intermittent special reflection or critic tokens (e.g., \emph{is\_supported}, \emph{is\_relevant}, etc.), leveraging knowledge distillation from proprietary models like GPT-4. At inference, these generated tokens determine the usability of each candidate output. While these methods enable the model to effectively rank candidate outputs from different retrievals and partially improve grounded 
generation, they struggle with navigating irrelevant or misleading information, especially when dealing with complex queries such as multi-hop retrieval tasks. This limitation arises since the models are not explicitly trained to contrast harder distractor passages and adhere to the facts from the retrievals. 

To confront the challenge, our framework \tool transforms an arbitrary dense LLM into a parameter-efficient (PEFT) sparse mixture of experts (MoE) model \cite{ wu2024parameter, komatsuzaki2022sparse} capable not only of self-reflection but also of handling complex reasoning tasks, including both single- and multi-hop queries. It uniquely trains the model to navigate challenging distractors that appear relevant but are misleading, while expanding the MoE only in the adapters, maintaining the model's scale. 
By combining constructive learning, architectural transformation, and reflection-based generation, \tool leverages latent learning, dynamically selects relevant experts, and integrates external knowledge effectively for more accurate and contextually supported response generation and estimates of their usefulness.

State-of-the-art (SoTA) open-LLM-based RAG models use external models to determine if retrieval is needed;  e.g., \citet{selfrag} use GPT-4 distillation and \citet{jeong2024adaptive-rag} use a finetuned FlanT5-XXL  for Llama2. However, since LLMs possess different parametric knowledge, it may not be effective to rely on another LLM to fully determine the retrieval necessity. To determine retrieval on-demand and balance performance and speed, we propose a hybrid adaptive retrieval method with two threshold alternatives based on model confidence. We train our model to generate \emph{retrieval/no\_retrieval} reflection tokens and measure the 
confidence 
of outputs conditioned on enforced \emph{no\_retrieval} during inference. If retrieval is needed, following \citet{selfrag}, we process all retrieved passages in parallel and rank them using the weighted linear sum of reflection token probabilities. Differently from other multi-step active or adaptive retrieval methods \cite{jeong2024adaptive-rag, jiang2023active, trivedi2023interleaving}, this eliminates the need for iterative generations.

\figmhtraindata
In experiments, we evaluate our framework on a wide range of single/multi-hop short/long-form knowledge-intensive reasoning tasks, including PopQA, TriviaQA, PubQA, Bio, ALCE-ASQA, HotpotQA, MuSiQue, and 2WikiMultiHopQA benchmarks.  Results show that our \tool significantly improves the overall factual accuracy and reasoning capabilities w.r.t the prior open-source RAG models, often matching or outperforming state-of-the-art proprietary LLMs and their RAG models. In multiple tasks, \toolnospace, based on Llama2-7B, sets new benchmarks, surpassing ChatGPT-RAG, Self-RAG, RAG 2.0, and 104B RAG-Command R+. Through detailed ablations, examples, and analysis, we provide further insights into the effectiveness of \toolnospace.

\section{\toolnospace: Enhanced Retrieval-Augmented Reasoning}

\tool transforms an arbitrary dense LLM into a parameter-efficient sparse MoE model capable not only of self-reflection but also of handling complex reasoning tasks.

Additionally, we devise an adaptive hybrid retrieval schema to balance the retrieval frequency and speed trade-off.
Below we first present the overview of \tool and then discuss 
the training, including dataset and fine-tuning, 
and hybrid adaptive inference.

\subsection{Overview}
\label{sec:overview}

We define \tool LLM as a model \llm~ that, given an input query \( q \)\footnote{With additional contexts if provided}, generates an output sequence of $m$ tokens \( o = [o_1, o_2, ..., o_m] \). To control model behavior and generate more context-supported responses, we adopt the reflection-based generation from Self-RAG \cite{selfrag} and augment output vocabularies with four types of special \emph{reflection} tokens: \emph{Retrieval}, \emph{Relevance}, \emph{Grounding} and \emph{Utility}. During training, given \( q \), the model learns to first generate the \emph{Retrieval} tokens (\texttt{[RT]/[NoRT]}) that indicate whether retrieval is necessary to answer $q$.\footnote{For long-form generation, we also use the \texttt{[Continue]} token, which indicates that the model can continue to use information from the previous segment.}
During inference, we employ a hybrid adaptive retrieval schema, leveraging both the \emph{Retrieval} tokens and model confidence. 

If no retrieval is needed, \llm~ generates the response using only the parametric knowledge of the LLM (i.e., return $o$ as $y_{pred}$). If retrieval is needed, for both single- or multiple-hop from an external knowledge source \( D = \{d_i\}_{i=1}^{N_d} \), we use a user-defined frozen retriever $R$ to retrieve the top-\( k \) documents \( S = \{s_t\}_{t=1}^{k} \), where each \( s_t \) consists of \( \{r_j\}_{j=1}^{N_H} \) with \( r_j \in D \) and \( N_H \) denoting the hop size. For each retrieved content $s_t$, \llm~ generates a \emph{Relevance} token, the output response \( y_t \), a \emph{Grounding} token, and a \emph{Utility} token.
The \emph{Relevance} tokens (\texttt{[Relevant/Irrelevant]}) indicate if \( s_t \) is relevant to \( q \), the \emph{Grounding} tokens (\texttt{[Fully Supported/Partially Supported/No Support]}) indicate if \( y_t \) is supported by \( s_t \), and the \emph{Utility} tokens (\texttt{[U:1]}-\texttt{[U:5]}) define how useful \( y_t \) is to \( q \). We process each  \( s_t \) in parallel and generate the final answer \( y_{pred} \) by ranking them (i.e., all $y_t$) based on the weighted sum of the normalized confidence of the corresponding predicted \emph{Relevance}, \emph{Grounding}, and \emph{Utility} tokens\footnote{For long-form generation, we use the same segment-level beam search strategy as in Self-RAG \cite{selfrag} to obtain the \texttt{Top-}$B$ segments, where \( B \) is the beam size, and return the best sequence at the end of generation.} (see Figure \ref{fig:inference-pipeline}).

\subsection{\tool Training}
\label{sec:open-rag-training}

Here, we discuss our training data collection (Sec \ref{sec:data-collection}) and 
  parameter-efficient
  MoE fine-tuning  (Sec \ref{sec:moe-finetuning}).

\subsubsection{Data Collection}
\label{sec:data-collection}

To empower \tool to tackle retrieval-free queries, as well as single- and multi-hop queries that require retrieval, we build our training data using various types of tasks and datasets. Given an input-output data pair ($q$, $y$) in an original dataset, we augment the data with \emph{reflection} tokens (Sec. \ref{sec:overview}) leveraging ground truth annotation or critic LLM  \(C\) to create supervised data. If the corresponding \emph{Retrieval} token added by \(C\) is \texttt{[RT]}, we further augment the data and create three different new instances accordingly as follows. First, we use $R$ to retrieve the top-$k$ documents \(S\). For each retrieved document \(s_t\), \(C\) evaluates whether \(s_t\) is relevant or not and returns the \emph{Relevance} token. To address both single- and multi-hop queries, we equip our data pipeline with a hop-unified heuristic: if at least one passage \( \{r_j\} \in s_t \) is relevant, we add the \emph{Relevance} token as \texttt{[Relevant]}; otherwise, we use \texttt{[Irrelevant]}. When \texttt{[Relevant]} is predicted, to enable \llm~ to contrast between useful and distractor contexts in \(s_t\) in a more fine-grained way, we design a data-contrastive heuristic: (i) for single-hop RAG datasets, we use \(C\) directly to label the \emph{Grounding} token; (ii) for multi-hop RAG datasets, if all passages \( \{r_j\} \in s_t \) are individually predicted as \texttt{[RT]}, then we add \texttt{[Fully Supported]} as the \emph{Grounding} token; otherwise, we use \texttt{[Partially Supported]}. Finally, regardless of the prediction of the \emph{Relevance} token, we use \(C\) to provide a \emph{Utility} score for \(y\) with respect to \(q\). We depict an example of the training data collection for a 2-hop question in Figure \ref{fig:mh_data}.

\subsubsection{Parameter-Efficient MoE Finetuning}
\label{sec:moe-finetuning}
RAG tasks are inherently complex, composed of various components such as queries with single (single-hop) or multiple (multi-hop) passages. The ability to leverage different parts of the model selectively based on such complexities can facilitate more adaptive and fine-grained reasoning capabilities over versatile input contexts. Therefore, instead of traditional dense models that treat all parts uniformly, we propose to transform \llm~ into a MoE architecture on the fly, which learns to selectively activate the most suitable experts dynamically for each query with versatile complexity (e.g., single/multi-hop). This selective activation is learned (fine-tuned) using our tailored training data, ensuring that the model learns to differentiate between useful and misleading information.

As open-source models are often used in low-compute settings, \tool employs sparse upcycling \citep{komatsuzaki2022sparse, wu2024parameter} to transform \llm~ into a parameter-efficient sparse MoE. This approach adds only a few million adapter parameters, preserving the same order of active parameters as in the original LLM.
\figmoearchitecture
The sparse MoE \tool model augments the FFN layer of the dense backbone LLM with a parameter-efficient MoE transformer block consisting of a set of expert layers $\mathbf{E} = \{\mathcal{E}_e\}_{e=1}^{N_E}$ along with an efficient routing mechanism as in Figure~\ref{fig:moe-architecture}. Each expert layer comprises a replicated original shared FFN layer weight, adapted by an adapter module $\mathcal{A}_{e}$ with parameters $\theta_e$. To ensure parameter efficiency, in each expert, we keep the FFN layer frozen and train the adapter module $\mathcal{A}_{e}$ only. In this way, we are only required to store one FFN replica keeping the model size unchanged except for the increase in the parameters in the adapter and the router modules. The rest of the layers, such as Norm and Attention, are copied from the dense model.

For a given input $x$, the router module $\mathcal{R}$ activates $\texttt{Top-}k$ experts out of $N_E$ experts based on the normalized output $x_{in}$ of the attention layer. Given $W_{|\cdot|}$ denotes the weight of the corresponding expert module, we define the router module as follows:
\begin{equation}
	\mathcal{R}(x_{in}) = \text{Softmax}(\texttt{Top-}k(W_{\mathcal{R}} \cdot x_{in}))
\end{equation} 

We formulate the adapter $\mathcal{A}_{e}$ as:
\begin{equation}
	\mathcal{A}_{e}(x) = \sigma(x W_{e}^{down}){W_{e}^{up}} + x.
\end{equation}

The efficiency of  \tool model results from the setup that 
% $|\theta| = \sum_{e=1}^{N_E} |W_{e}^{down}| + |W_{e}^{up}|  \ll |\phi_o|$ 
$|\theta_e| = |W_{e}^{down}| + |W_{e}^{up}|  \ll |\phi_o|$ 
where we keep $\phi_o$ from the 
dense LLM frozen during fine-tuning. Finally, we express the output $y$ of a parameter-efficient expert module as:
\begin{equation}
	y = \sum_{e=1}^{N_E} \mathcal{R}(x)_e \mathcal{A}_e (\mathcal{E}_e(x)).
\end{equation}

In our implementation, we use $N_E = 8$ and $k = 2$ if not otherwise specified. In other words, only $2$ of the $8$ experts are active during training and inference. We train \tool with QLoRA~\cite{dettmers2023qlora} adapters during fine-tuning which has a 
load-balancing objective along with the standard conditional language modeling objective. To mitigate the approximation error in the expert adapters, we use the adapters with a dimension of $512$ by default.

\subsection{Hybrid Approach for Adaptive Retrieval}
\label{sec:adaptive_inference}

Since LLMs possess different parametric knowledge, instead of using other LLMs, we propose a hybrid adaptive retrieval method with two threshold alternatives based on model confidence to determine retrieval on-demand and balance performance speed. We take motivation from both control token-based ~\citep{selfrag, lu2022quark} and confidence-based ~\cite{liu2023litcab, jiang2023active} inference methods. 

 During training, \llm~ learns to generate \emph{Retrieval} reflection tokens (\texttt{[RT]} and \texttt{[NoRT]}). At inference, we measure the confidence of the output sequence \( o \) conditioned on an enforced no retrieval setting by adding \texttt{[NoRT]} to the input, such that \(\hat{q} = q \oplus \texttt{[NoRT]}\). We design two different confidence scores \( f_{|\cdot|} \): (i) \( f_{minp} \), the minimum value of the probabilities of the individual tokens, and (ii) \( f_{meanp} \), the geometric mean of the probabilities of the individual tokens in the generated sequence.
\begin{align}
\label{eq:meanpscore}
f_{minp}(o | \hat{q}) &= \min_{i=1}^{m} p(o_i|\hat{q}, o_{<i}) \\
f_{meanp}(o| \hat{q}) &= \sqrt[m]{\prod_{i=1}^{m} p(o_i|\hat{q}, o_{<i})} 
\end{align}
We control retrieval frequency with a tunable threshold \( \gamma \), where retrieval occurs if \( f_{|\cdot|}<\gamma \).

\section{Experiments}
\label{sec:exp}

\subsection{Tasks and Datasets}
\label{sec:sub:task-daatset}

\minisection{Single-hop short-form tasks} include PopQA \citep{mallen2022not}, TriviaQA-unfiltered \citep{joshi-etal-2017-triviaqa}, and PubHealth \citep{zhang2023interpretable}. These datasets involve answering factual questions and verifying public health facts, using retrieved contexts provided by Self-RAG. We use the accuracy metric for evaluation.

\minisection{Single-hop long-form generation tasks} cover biography generation (Bio) \citep{min2023factscore} and the long-form QA benchmark ALCE-ASQA \citep{gao2023enabling,stelmakh2022asqa}. Biographies are evaluated with FactScore \citep{min2023factscore}, while ALCE-ASQA uses official metrics for correctness (str-em) and fluency based on MAUVE \citep{pillutla2021MAUVE}.

\minisection{Multi-hop reasoning tasks} include HotpotQA (distractor dev split) \citep{yang-etal-2018-hotpotqa}, MuSique-Ans \citep{musique}, and 2WikiMultihopQA \citep{DBLP:journals/corr/abs-2011-01060} which require systems to answer complex multi-hop questions. We use official EM and F1 metrics for evaluation.

\subsection{Experimental settings}

\minisection{Training Data and Settings.}
In our data curation process, as detailed in Section \ref{sec:data-collection}, we compile a diverse set of instruction-following input-output pairs encompassing retrieval-free, single-hop, and multi-hop datasets requiring retrieval. For no-retrieval and single-hop datasets, we utilize 150K instruction-output pairs curated by Self-RAG.
For the multi-hop dataset, we randomly sample 16K two-hop instances from the HotpotQA \cite{hotpotqa} Distractor train split, each with 10 passages annotated with the ground truth \emph{Relevance} tokens. Using our data collection method from Section \ref{sec:data-collection}, we generate 28K new multi-hop training instances. All other \emph{reflection} tokens are labeled by the Llama2\modelsevenB~\cite{touvron2023llama} critic LLM in Self-RAG, which is distilled from GPT-4. Additional information regarding training is provided in Appendix Section \ref{sec:training}. Following previous works and for a fair comparison, we use the Llama2\modelsevenB~ \cite{touvron2023llama} as the base RAG model \llm.  \tool is transformed into a MoE model with $N_E = 8$ and $k = 2$, incorporating adapters with a dimension of $512$, totaling an additional (8$\times$135M) adapter model parameters. Moreover, we train a larger version of \tool based on Llama2\modelthirteenB~ with additional (8$\times$213M) parameters to demonstrate the scalability of our framework. By \tool model, we indicate  \toolnospace\modelmoesevenB~ if not explicitly mentioned.

\minisection{Inference Data and Settings.}
We assign the default weight of 1.0, 1.0, and 0.5 to \emph{Relevance}, \emph{Grounding}, and \emph{Utility} tokens respectively. Following Self-RAG,  
we compare the model performances with always retrieval and vary the retrieval frequency as discussed in Sec \ref{sec:adaptive_inference} only to demonstrate optimum thresholding and performance-speed trade-offs. In multi-hop evaluations, from the corresponding retrieval candidate passages, we use Beam Retriever \cite{beamret} to retrieve \texttt{Top}-$3$ multi-hop contexts, each with the mentioned $N_H$ number of passages. For single-hop tasks, we use Self-RAG's setup (See Appendix ~\ref{sec:inference}).

\tabmain

\subsection{Baselines}
\minisection{Baselines without retrievals.}
We compare ours with several strong, publicly available pre-trained LLMs, including Llama2-7B,13B ~\citep{touvron2023llama}, SAIL-7B~\citep{luo2023sail} as well as instruction-tuned models, Alpaca-7B,13B \citep{dubois2023alpacafarm}. Additionally, we consider models trained and reinforced with private data such as ChatGPT \citep{ouyang2022training}. For instruction-tuned LMs, we utilize the official system prompt or instruction format of the corresponding model.

\minisection{Baselines with retrievals.}
We evaluate models incorporating retrieval during both testing and training phases, focusing on standard Retrieval-Augmented Generation (RAG) baselines with open-source Large Language Models (LLMs) like Llama2, Alpaca and LongChat~\citep{longchat2023}. These models generate outputs based on queries alongside top retrieved documents using our retriever. We also present results for RAG baselines utilizing private data, including RAG-ChatGPT, RAG2.0~\citep{rag2point0}, and RAG-Command R+~\citep{commandrplus}, which prepend top-retrieved documents to the query. 
Additionally, we assess RQ-RAG~\citep{chan2024rq}, which employs proprietary retriever models.
Finally, our comparisons extend to Perplexity.ai, Self-RAG~\citep{selfrag}, and SAIL~\citep{luo2023sail}, which are also finetuned with retrieved texts.

\figthreshold

\section{Results and Analysis}
Here, we (i) evaluate the RAG models (ii) demonstrate the effectiveness of our adaptive retrieval in balancing the performance-speed (iii) present ablation studies and further analysis.

\subsection{Main Results}
\label{sec:main-result}

\textbf{Comparison against baselines without retrieval.}
Table \ref{tab:main} (top and middle blocks) shows the performance of open-source baselines without retrieval. \tool demonstrates substantial performance gains over all supervised fine-tuned LLMs, many of which are larger in size (e.g., 65B CoVE) and even our \tool outperforms ChatGPT across all metrics and tasks. Particularly in multi-hop reasoning tasks such as HotpotQA, \tool achieves a significant EM score of 63.3\%, surpassing Alpaca\modelthirteenB's 0.7\%. In contrast, while ChatGPT achieves a decent score of 22.4\% EM in HotpotQA, its performance drops notably in other multi-hop tasks like MuSiQue, where it achieves only 3.1\% EM while \tool achieves a much higher score of 41.6\% EM in MuSiQue, highlighting its robustness and effectiveness in complex query handling compared to both open-source and proprietary LLMs.

\textbf{Comparison against baselines with retrieval.} As shown in Table \ref{tab:main} (bottom), \tool consistently outperforms existing open-source RAG models, even those larger in size. It achieves the top performance among non-proprietary LM-based models across all tasks, with the exception of TriviaQA and PubQA, where it is marginally surpassed (by 1.2\% and 0.4\%, respectively) by the larger Self-RAG\modelthirteenB~model, and by Alpaca\modelthirteenB~in a single metric within the ALCE-ASQA dataset. 

We observe that while baseline open-source RAG models achieve higher accuracy, even surpassing strong proprietary models like RAG-ChatGPT in single-hop reasoning tasks, their performance significantly lags in multi-hop reasoning tasks. Our contrastive learning of the distractor contexts substantially enhances the reasoning in \tool and empowers it to outperform the proprietary RAG-ChatGPT in all complex multi-hop datasets. 

Moreover, \tool surpasses RAG 2.0 and 104B Command R+, which are specifically built for RAG tasks, in HotpotQA (63.3\% vs. 60.0\% EM) and PubQA (75.9\% vs. 46.3\% Acc). In long-form generation, proprietary models often achieve higher scores, but ours remains highly competitive. For instance, RAG-Command R+ attains a FactScore (FS) of 84.0\% in Bio, slightly outperforming \toolnospace's 82.2\%. 
In addition, our \toolnospace\modelmoethirteenB~model outperforms all baselines in all multi-hop tasks; and all open baselines in all short-form tasks and shows competitive performance with the proprietary models. 
These results highlight the superior ability of \tool to effectively integrate and utilize retrieved information, enhancing both reasoning accuracy and fluency across varying complexities and both short- and long-form generations.

\subsection{Performance-Speed by Adaptive Retrieval }
\label{sec:adaptive_inference_results}
As discussed in Sec \ref{sec:adaptive_inference}, given the query, adaptive retrieval method provides a probability/confidence score from the model. By thresholding on that score, we can control the retrieval frequency and balance the performance-speed trade-off and this can also guide to determine when retrieval is needed.  A better scoring method should achieve higher accuracy at any retrieval frequency. 
In order to demonstrate our hybrid adaptive retrieval scoring over the existing reflection token probability-based method $f_{ret}$ in Self-RAG, in Figure~\ref{fig:tradeoff}, we plot the downstream accuracy vs retrieval frequency (top), and accuracy vs confidence score (bottom) for PopQA, PubHealth, and TriviaQA datasets by sweeping across different threshold values $\gamma$ (larger $\gamma$ causes less retrieval) from 0 to 1. In Figure~\ref{fig:tradeoff} (bottom), we notice that for $f_{meanp}$ or $f_{minp}$, the accuracy increases with higher values of confidence while $f_{meanp}$ is more robust, showing monotonically increasing accuracy with higher confidence scores consistently in all dataset.  But in the case of $f_{ret}$, no such pattern exists. Overall (top) as these benchmarks are knowledge-intensive, they typically perform better with retrieved contexts and our adaptive scoring shows a better determination of when to retrieve and when not -- resulting in higher accuracy at any retrieval frequency. In fact, the advantage is more amplified in PubHealth where we can find a clear threshold confidence score which if achieved, retrieval data are found to be less effective than the parametric knowledge. This gives us a peak accuracy of 1\% more than always retrieval, which can not be determined by Self-RAG.

\subsection{Ablation Studies}
\label{sec:ablation}
\figcrag
\figrouteall

\vspace{12pt}
\minisection{Robustness to Different Retrieval (CRAG) Methods.}
CRAG~\cite{yan2024correctivecrag} proposes a corrective RAG method where, if corpus (e.g., Wikipedia) retrievals are detected as low-quality, a web search is performed to obtain new retrievals. These new retrievals are then fed into the system. The Self-CRAG method combines both reflection-based models and CRAG-based datasets (Self-RAG + CRAG dataset). We evaluate \tool and \opencrag (\tool + CRAG datasets) on the benchmarks (PopQA, PubHealth, and Bio) using CRAG, Self-RAG~\cite{selfrag}, and Self-CRAG as baselines, as illustrated in Figure~\ref{fig:crag}. \opencrag outperforms all baselines across all tasks. Specifically, \tool achieves 2\%, 4\% higher accuracy than Self-CRAG in (Bio, PopQA) and PubHealth respectively. This demonstrates \toolnospace's robustness to retrieval quality and its potential for improvement with high-quality contexts.

\figmoedensesparse
\minisection{Routing Analysis of \toolnospace.}
We perform routing analysis for PopQA, PubHealth, HotpotQA, and 2WikiMultihopQA tasks to demonstrate \texttt{Top-}$2$ expert activation in different layers during retrieval-free generation by \tool as illustrated in Figure~\ref{fig:route_all}. 
We observe, that $\mathcal{E}_7$ is a general expert that is highly activated in the first (Layer 1), middle (Layer 16), and final (Layer 32) layers for all datasets. Whereas $\mathcal{E}_2$ is activated in the first layer while $\mathcal{E}_6$ is activated mostly in the final layer. In the middle layer, we also observe a higher activation of $\mathcal{E}_5$ and a lower activation of $\mathcal{E}_7$ in the PopQA and PubHealth datasets (single-hop), but the opposite in the case of multi-hop datasets -- showing that the experts implicitly learn to identify query complexity and play important roles across layers for different kinds of task complexities.
 
\minisection{Sparse Upcycling Hyperparameters.} We experiment with different hyper-parameters of \tool  as shown in Table~\ref{tab:moe_hp}. We observe that increasing the number of experts $N_E$ slightly improves the performance in MuSiQue, and performance improvement in training longer (epoch 1 vs 2). Increasing the number of active experts $k$ from 2 to 4 causes performance degradation showing the necessity of less active experts. 

\minisection{Impact of Modules.} 
It is important to understand how much gain is coming from our contrastive learning and how much from the architectural transformation. 
In Figure~\ref{fig:dense-moe} with reference to Self-RAG, we plot 
 \tool performances with both dense and MoE architecture. 
\toolnospace-Dense outperforms Self-RAG-7B by 1.8\% in PopQA, 1.6\% in PubHealth, 4.2\% in ASQA (MAUVE), 17.9\% in MuSiQue (EM) and 21.7\% in HotpotQA (EM). 
Moreover, \toolnospace-MoE improves over \toolnospace-Dense by 1.6\% in PopQA, 2.2\% in PubHealth, 5.2\% in ASQA (MAUVE), 1.6\% in MuSiQue (EM) and 1.4\% in HotpotQA (EM) -- both components enhances the model significantly while contrastive learning as highest.

\tabmoeablation

\section{Related work}
\label{sec:related-work}

Complex factual reasoning requires contextualizing information from multiple documents~\cite{musique, hotpotqa}. Prior works~\cite{dsp, self-ask, DBLP:conf/ecir/PereiraFLN23, decomp} proposed decomposing multi-hop queries into single-hop queries, then repeatedly using LLMs and Retrievers. In addition, ~\citet{ActiveRetrieval}  retrieved new documents if the tokens within generated sentences have low confidence. However, the performance improvement of these approaches often comes at the cost of resource-intensive techniques such as interleave Chain-of-Thought~\cite{react, ircot, zhang2024raft} or Tree-of-Thought~\cite{chan2024rq} reasoning with document retrieval; and requiring external models ~\cite{jeong2024adaptive-rag}. In this work, we train a single MoE model capable of answering complex questions in one iteration with a minimal increase in model complexity.

\section{Conclusion}
\label{sec:conclusion}
To enhance reasoning capabilities in RAG models with open-source LLMs, we develop \tool featuring a PEFT MoE architecture, contrastive learning, and adaptive retrieval. 
\tool shows significant performance improvements in complex reasoning tasks, outperforming SoTA methods. However, there is still a gap in tasks like long-form generation compared to proprietary models, which we aim to address in future.

\section{Limitations}

\tool has a higher memory footprint due to an increase in total parameters (7.81B) in comparison to Llama2\modelsevenB~family baselines (6.74B). But our \tool outperforms open LLMs with total parameters ranging from 7B to 65B, rivaling proprietary models such as ChatGPT, Perplexity.ai, and Command R+ in various downstream tasks. Thus, \tool eventually reduces the compute and memory cost with 7.01B active parameters during inference in comparison to its performance. Additionally, as our framework is general, future direction can be building stronger sparse-upcycled LLMs based on recent models such as Llama3\modeleightB~ and Mistral\modelsevenB~ utilizing \tool multi-hop training dataset. Although our approach is theoretically applicable to any domain, future work can explore developing high-performance domain-specific RAG based on our \toolnospace.

\section*{Acknowledgement}

We thank anonymous reviewers for their valuable feedback on the paper.
We also thank Mohamed El Banani and Amr Keleg for fruitful discussions.
We are grateful to Qatar Computing Research Institute for providing compute and OpenAI APIs.
Shayekh Bin Islam is supported by the Fatima Al-Fihri Predoctoral Fellowship sponsored by Hugging Face.
This work was supported in part by National Science Foundation (NSF) awards CNS-1730158, ACI-1540112, ACI-1541349, OAC-1826967, OAC-2112167, CNS-2100237, CNS-2120019, the University of California Office of the President, and the University of California San Diego's California Institute for Telecommunications and Information Technology/Qualcomm Institute. Thanks to CENIC for the 100Gbps networks.

\bibliography{ref.bib}

% \newpage
% \clearpage
\newpage

\clearpage

\appendix

\section{Training Details}
\label{sec:training}

We train both MoE and Dense models with LoRA rank 64, LoRA $\alpha$ 16, and LoRA dropout 0.1. We optimize the models with the AdamW optimizer with a linear learning rate scheduler and a weight decay of 0.0. Both models have a context length of 4096 for facilitating long-context multi-hop QAs.
Other training hyper-parameters are mentioned in Table~\ref{tab:training_hps}.

\begin{table}[h]
    \centering
    \resizebox{0.5\textwidth}{!}{
    \begin{tabular}{c | cccc} 
        \toprule
          LM & LR & Epoch & Quantization & Adapter Dim \\
          \midrule
          Dense\modelsevenB & $1 \times 10^{-4}$ & 3 & None & -- \\
          MoE\modelsevenB & $2 \times 10^{-4}$ & 2 & QLoRA (NF4) & 512 \\
          MoE\modelthirteenB & $1 \times 10^{-4}$ & 2 & QLoRA (NF4) & 512 \\
        \bottomrule
    \end{tabular}
    }
    \caption{Training Hyper-parameters.}\label{tab:training_hps}
\end{table}

We train \tool models using NVIDIA A100 GPUs with 80GB VRAM. 
About 40 GPU days have been spent in total during training and model development.

\subsection{Dataset Details}
The complete breakdown of \tool training dataset is displayed in Table~\ref{tab:mgen_training}. Algorithm~\ref{alg:mh_train_data_psuedocode} shows the process of the multi-hop training data preparation.

\begin{table}[ht!]
\centering
\small
\begin{tabular}{l c c}
\toprule
Dataset Name & Source & Number of Instances  \\
\midrule

\multicolumn{3}{c}{\it Instruction-Following} \\  
\midrule
GPT-4 Alpaca & Open-Instruct & 26,168\\ 
Stanford Alpaca & Open-Instruct & 25,153 \\ 
FLAN-V2 & Open-Instruct &17,817 \\
ShareGPT & Open-Instruct & 13,406 \\
Open Assistant 1 & Open-Instruct & 9,464 \\
\midrule
\multicolumn{3}{c}{\it Knowledge-Intensive (Single-Hop)} \\
\midrule
Wizard of Wikipedia &   KILT & 17,367  \\
Natural Questions & KILT & 15,535\\ 
FEVER &  KILT & 9,966\\ 
OpenBoookQA &  HF Dataset & 4,699 \\ 
Arc-Easy &   HF Dataset & 2,147 \\ 
ASQA &  ASQA &3,897 \\ 

\midrule
\multicolumn{3}{c}{\it Knowledge-Intensive (Multi-Hop)} \\
\midrule
\rowcolor[gray]{0.9} HotpotQA (Ours) & HotpotQA &  28,117 \\
\bottomrule
\end{tabular}
\caption{The generator LM training data statistics. Instruction-following and single-hop knowledge-intensive samples are from Self-RAG~\cite{selfrag}. We curate the multi-hop knowledge-intensive samples with reflection tokens.
}\label{tab:mgen_training}
\end{table}

\section{Inference Details}
\label{sec:inference}

\subsection{Inference Hyper-parameters}

The weights of the \reltok, \gndtok~ and \utiltok~ tokens types are 1.0, 1.0, and 0.5 respectively during inference of \tool and Self-RAG. During long-form generation, we use the maximum depth of search of 7 and the size of the beam of 2 following Self-RAG. To evaluate the performance in the retrieval setting, we report the performance in the always retrieval setup in Table~\ref{tab:main}. 
Next, we employ greedy decoding for \tool and Self-RAG; and top-$p$ (nucleus) sampling for open baseline models with temperature 0.8 and $p = 0.95$.

We discuss the different soft retrieval constraints in Section~\ref{sec:adaptive_inference} and Section~\ref{sec:adaptive_inference_results}. 
Moreover, we identify a bug \footnote{\href{https://github.com/AkariAsai/self-rag/blob/1fcdc420e48f50a7d7ab1ece5494221b93252e99/retrieval_lm/run_short_form.py\#L79}{Implementation issue of soft-constraint in Self-RAG}}
in the implementation of soft-constraint for adaptive retrieval in Self-RAG where the implementation utilizes the log-probability of the \textit{Retrieval} token instead of the probability.

\subsection{Instruction Format}

We utilize standard prompt without any complex prompting, such as Chain-of-Thoughts (CoT). For single-hop tasks, we follow the instruction format in Self-RAG, whereas the instruction format for multi-hop question answering is shown in Table~\ref{tab:mh_instruction}.

\begin{table}[ht!]
\resizebox{0.5\textwidth}{!}{
    \begin{tcolorbox}
    {\bf Instructions}\\    
    \texttt{
    You are a question answering agent. Given a context and a question, your task is to answer the question based on the context. Instead of a full sentence, your answer must be the shortest word or phrase or named entity. Some example outputs 'answer' are: yes; no; Ibn Sina; Doha, Qatar; 2,132 seats, Los Angeles, California etc.}
    \tcblower
    \texttt{\#\#\# Instruction \\
    What administrative territorial entity is the owner of Ciudad Deportiva located?}
    \newline\newline
    \texttt{\#\#\# Response:}
    \newline
    \end{tcolorbox}
}
\caption{Instruction Example for Multi-Hop QAs.}\label{tab:mh_instruction}

\end{table}

\begin{algorithm*}[h]
\caption{\tool Multi-Hop Training Data Preparation}\label{alg:mh_train_data_psuedocode}
\begin{algorithmic}[1]
\Require Critic Model $C$, Multi-hop Reasoning QA collections $(Q, Y)$ with a set of supporting contexts $\mathcal{P}_i$ and a set of non-supporting contexts $\mathcal{N}_i$ for QA pair $(q_i, y_i)$.
\State {\bf Output:} 
Multi-hop input-output pairs $\hat{D}$.

\State $C$ predicts \textit{Retrieval} for $q_i$ and \textit{Utility} $U$ of $y_i$ for answering $q_i$.
\State Initialize an empty list $\hat{D}$

\For{$(q_i, y_i) \in \{Q, Y\}$} 
\If{ \textit{Retrieval} == \texttt{[NoRT]} }
    \State $\rho_0$ = 
    \texttt{[NoRT]} $\oplus\ y_i\ \oplus\ $ $U$
    \State $\hat{D} \coloneqq \hat{D} \cup \{ (q_i, \rho_0 )\}$
\ElsIf{ \textit{Retrieval} == \texttt{[RT]}}

    \State // Relevant and fully supported context
    
    \State Without replacement, uniformly sample two contexts $(p_i^1, p_i^2) \subseteq \mathcal{P}_i$

    \State $\rho_1$ = 
    \texttt{[RT]} $\oplus$ \texttt{<p>} $\oplus\ p_i^1 \oplus p_i^2\ \oplus$ \texttt{</p>} $\oplus$ \texttt{[Relevant]} $\oplus y_i\  \oplus$  {\footnotesize \texttt{[Fully supported]}} $\oplus\ $ $U$ 
    
    \State // Relevant and partially supported context
    
    \State Randomly sample one context $p_i^3 \in \mathcal{P}_i$
    
    \State Randomly sample one context $n_i^1 \in \mathcal{N}_i$
    
    \State $\rho_2$ = 
    \texttt{[RT]} $\oplus$ \texttt{<p>} $\oplus\  p_i^3 \ \oplus\  u_i^1\ \oplus$ \texttt{</p>} $\oplus$ \texttt{[Relevant]} $\oplus y_i\oplus$  {\footnotesize \texttt{[Partially supported]}} $\oplus\ $ $U$
    \State // Irrelevant context
    
    \State Without replacement, uniformly sample two contexts $(n_i^2, n_i^3) \subseteq \mathcal{N}_i$
    
    \State $\rho_3$ =
    \texttt{[RT]} $\oplus$ \texttt{<p>} $\oplus\  n_i^2 \ \oplus n_i^3\ \oplus$ \texttt{</p>} $\oplus$ \texttt{[Irrelevant]} $\oplus\  y_i\  \oplus$ $U$
    
    \State $\hat{D} \coloneqq \hat{D} \cup \{ (q_i, \rho_1), (q_i, \rho_2), (q_i, \rho_3) \}$ 

\EndIf

\EndFor

\end{algorithmic}
\end{algorithm*}

\end{document}